\documentclass{article}

 \usepackage[preprint]{neurips_2026}

\usepackage{graphicx}    
\usepackage{caption}     
\usepackage{array}       
\usepackage{booktabs}    
\usepackage{xcolor}
\usepackage{placeins}        
\usepackage{amsmath}         
\usepackage{amsfonts}        
\usepackage{amssymb}         
\usepackage{multirow}        
\usepackage{algorithm}
\usepackage{algpseudocode}  

\usepackage{booktabs}
\usepackage{ulem}
\usepackage[utf8]{inputenc} 
\usepackage[T1]{fontenc}    
\usepackage{hyperref}       
\usepackage{url}            
\usepackage{booktabs}       
\usepackage{amsfonts}       
\usepackage{nicefrac}       
\usepackage{microtype}      
\usepackage{xcolor}         

\usepackage{multirow}     
\usepackage{booktabs}     
\usepackage{graphicx}     
\usepackage[table]{xcolor} 
\usepackage{colortbl}       
\usepackage{graphicx}    
\usepackage{caption}     
\usepackage{array}       
\usepackage{booktabs}    
\usepackage{placeins}        
\usepackage{amsmath}         
\usepackage{amsfonts}        
\usepackage{amssymb}         
\usepackage{multirow}        

\usepackage{algorithm}      

\usepackage{arydshln}
\usepackage{enumitem}
\usepackage{wrapfig}
\usepackage{natbib}
\usepackage{wrapfig}  

\usepackage{amsmath}    
\usepackage{amssymb}    
\usepackage{amsthm}     
\usepackage{wrapfig}
\usepackage{booktabs}
\usepackage{multirow}
\usepackage{array}

\usepackage{algorithm}
\usepackage{algpseudocode}
\usepackage{xcolor}
\usepackage{amsmath,amssymb}
\usepackage[most]{tcolorbox}
\algnewcommand{\LeftComment}[1]{%
  \Statex \(\triangleright\) #1%
}

\usepackage{amsmath,amssymb,amsthm}

\newtheorem{theorem}{Theorem}[section]
\newtheorem{proposition}[theorem]{Proposition}
\newtheorem{corollary}[theorem]{Corollary}

\theoremstyle{remark}

\newtheorem{definition}{Definition}
\usepackage{amsthm}

\newcommand{\xu}[1]{#1}

\title{MaPP: A Unified Marginalized Posterior-Predictive Framework for Data-Efficient RLVR}

\author{%
  \textbf{Yangyang Ren}$^{1,2}$\thanks{Equal contribution.} \qquad
  \textbf{Haodong Zhu}$^{1,2}$\footnotemark[1] \qquad
  \textbf{Sheng Xu}$^{3}$\thanks{Corresponding authors:
    Sheng Xu (\texttt{shengxu@cuc.edu.cn}) and
    Yanjing Li (\texttt{yanjing.li@ntu.edu.sg}).} \qquad
  \textbf{Yanjing Li}$^{4}$\footnotemark[2] \\
  \textbf{Nikolai Yu. Zolotykh}$^{5}$ \quad
  \textbf{Wentao Zhang}$^{2,6}$ \quad
  \textbf{Baochang Zhang}$^{1,7}$ \\[0.5em]
  $^{1}$Beihang University \quad
  $^{2}$Zhongguancun Academy \quad
  $^{3}$Communication University of China \\
  $^{4}$Nanyang Technological University \quad
  $^{5}$Lobachebsky University \\
  $^{6}$Peking University \quad
  $^{7}$Hangzhou Innovation Institute of Beihang University
}

\begin{document}

\maketitle

\begin{abstract}
Reinforcement learning with verifiable rewards (RLVR) enhances the reasoning capabilities of large language models (LLMs), but at the expense of significant computational overhead due to compute-intensive rollout processes and frequent policy updates. 
Online prompt selection, a \xu{widely adopted} strategy for improving training efficiency, maintains per-prompt Bayesian posteriors to predict prompt difficulty and prioritize informative prompts before committing rollout budget.
\xu{However, these methods assess prompt informativeness without accounting for how reliably learning signal is extracted from sampled responses. In GRPO-based RL algorithms, the realized advantage of a response depends not only on its own outcome, but also on the randomly sampled outcomes of its peers through group normalization.}
%
\xu{Our experimental and theoretical analysis 
show that the resulting uncertainty in group composition induces \textit{composition noise}, a non-vanishing variance component that creates an irreducible lower bound on gradient estimation error.
%
Consequently, the standard group-relative advantage fails to faithfully characterize response-level utility, degrading gradient estimation and thereby impairing downstream prompt selection.}
To address this issue, we propose a unified \emph{\textbf{Ma}rginalized \textbf{P}osterior-\textbf{P}redictive} framework (\emph{\textbf{MaPP}}) for data-efficient RLVR, \xu{which first denoises response-level advantage estimation and then improves prompt selection using a shared Beta posterior.}
\xu{Specifically, for each response, MaPP replaces the standard group-relative advantage with a composition-invariant intrinsic advantage via closed-form Beta–Binomial marginalization.}
This yields a closed-form posterior-predictive advantage estimator whose error provably diminishes as the posterior concentrates.
\xu{Then, building on the same posterior, MaPP derives an uncertainty-aware prompt selection score that more faithfully characterizes prompt informativeness, improving data efficiency without additional rollout cost.}
Experiments across mathematics, planning, and visual geometry on five model backbones show that MaPP consistently outperforms GRPO and strong selection baselines, \xu{achieving up to +2.45 average accuracy improvement over the strongest baseline under the same rollout budget, a new state-of-the-art.}

\end{abstract}

\section{Introduction}
\label{sec:intro}


Reinforcement learning with verifiable rewards (RLVR) is a key post-training paradigm for enhancing LLM reasoning~\citep{deepseekmath,dapo,deepseek,jaech2024openai,yang2025qwen3}. Among RLVR algorithms, GRPO~\citep{deepseekmath} is most widely adopted, eliminating a learned value network by estimating advantages via within-group normalization of G responses per prompt. However, RLVR incurs high computation and memory costs due to intensive rollouts for policy evaluation and updates~\citep{zheng2025act,lin2025cppo}.
\xu{Thus, improving utility per fixed rollout budget is a central challenge in scaling RLVR.}

\xu{Existing methods improve training efficiency by selecting prompts for rollout budget.}
Uniform sampling is inefficient: overly easy or hard prompts cause degenerate groups (all correct or incorrect), wasting rollouts on zero gradient~\citep{bae2026online,chen2025self,zeng2025cures}. Only intermediate-difficulty prompts provide informative signals. Dynamic Sampling~\citep{dapo} oversamples a candidate set, rolls out all, then discards degenerate ones—but the rollouts on discarded prompts are already spent~\citep{zheng2025act}.

\begin{figure*}[t]
    \centering
    \includegraphics[width=\textwidth]{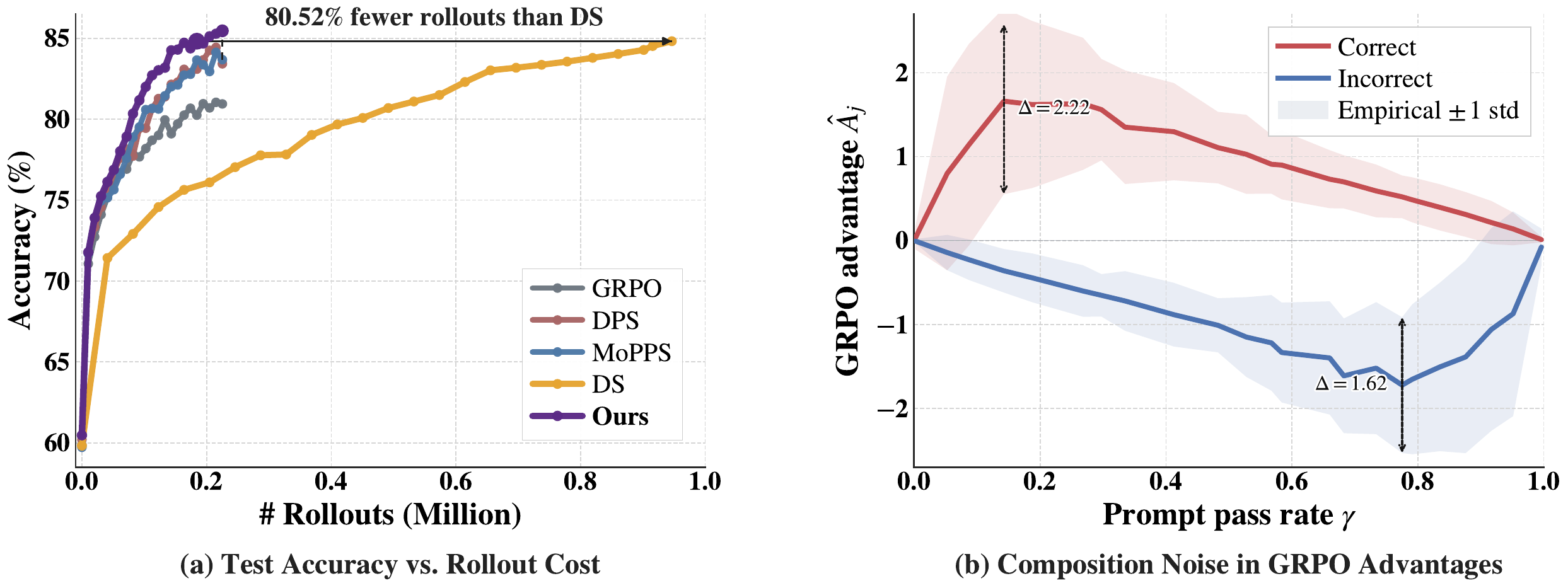}
    \caption{On the CountDown dataset using the Qwen3-4B model, {(a)} MaPP achieves higher accuracy than
    all selection methods with the same rollout cost, while requiring 80\% fewer rollouts than DS to reach
    comparable performance.
    {(b)} Group-composition uncertainty induces \emph{composition noise} in GRPO (group size $G{=}8$; mid-training checkpoint of MoPPS; 100 prompts $\times$ 50 groups): under the same policy checkpoint, identical correct (red) or incorrect (blue) response receives widely varying advantages depending on group composition ($\pm 1$ std shaded). The noise persists across all non-degenerate difficulty levels, {\em e.g.}, \ $\Delta=2.22$ at
    $\gamma\approx0.15$ and $\Delta=1.62$ at
    $\gamma\approx0.8$.}
    \vspace{-25pt}
\label{fig:teaser}
\end{figure*}

To avoid this waste, prediction-based methods~\citep{mopps,greso,insight} estimate prompt difficulty \emph{before} rollout and skip likely degenerate prompts entirely, achieving comparable or superior performance with significantly fewer rollouts, \xu{as shown in Fig.~\ref{fig:teaser} (a).} 
%
\xu{However, conventional methods which select prompt before rollout do not by itself determine how reliably learning signal is extracted from the responses sampled afterward.}
%
%
\xu{In GRPO-based RLVR~\citep{deepseekmath}, both prompt selection and policy optimization are driven by group-level statistics formed from the sampled responses. For non-degenerate prompts, these counts are not fixed properties of the prompt itself, but random rewards with uncertainty induced by the current policy's latent success rate under a finite number of sampled responses. As a result, even for the same response on the same prompt under the same policy, the realized advantage can deviate substantially depending on which peers happen to be drawn.}
\xu{Fig.~\ref{fig:teaser} (b) visualizes this effect: the same correct or incorrect response receives widely varying advantages across different sampled groups, despite sharing the same prompt and policy.}
\xu{We refer to this deviation, induced by uncertainty in group composition, as \emph{composition noise}.}
\xu{This composition noise degrades gradient estimation and, in turn, limits how faithfully prompt informativeness can be assessed for downstream selection.}
%

\xu{Crucially, the information needed to account for this noise already exists in current selection pipelines: the per-prompt Beta posterior maintained over the latent pass rate $\gamma_\tau$ tracks exactly the quantity that governs group-composition uncertainty.}
To leverage this observation, we propose a \emph{\textbf{Ma}rginalized \textbf{P}osterior-\textbf{P}redictive} framework (\emph{\textbf{MaPP}}), \xu{which first denoises response-level advantage estimation and then improves prompt selection using a shared Beta posterior.}
The key insight is that the GRPO advantage conditions on a single realized group composition, whereas the \emph{true} learning signal of a response is determined by its correctness and the prompt's intrinsic difficulty alone. 
The first component of MaPP, which we call \textbf{MaPP-AD}, denoises the realized GRPO advantage via closed-form Beta--Binomial marginalization, recovering a composition-invariant \emph{intrinsic advantage}. 
This yields a closed-form posterior-predictive advantage estimator whose error provably diminishes as the posterior concentrates.
\xu{Building on the same posterior, the second component of MaPP, which we call \textbf{MaPP-PS}, derives a marginalized posterior-predictive, uncertainty-aware prompt selection method that more accurately characterizes prompt informativeness. Both components are modular and can be layered on top of existing selection pipelines.}
Our contributions are threefold:
\begin{itemize}[leftmargin=*, itemsep=2pt, topsep=2pt]
    \item We identify \emph{composition noise}, a within-rollout variance component in the GRPO advantage \xu{induced by uncertainty in group composition} and show that it degrades gradient estimation and thereby limits downstream prompt selection (\S\ref{sec:motivation}).
    \item We propose MaPP, a unified posterior-predictive framework \xu{with two closed-form components: {MaPP-AD}, which denoises the realized GRPO advantage via posterior-predictive marginalization, and {MaPP-PS}, which uses the same shared Beta posterior to derive a more faithful prompt-selection score (\S\ref{sec:method}).}
    \item Experiments across three domains and five model backbones show that MaPP consistently outperforms GRPO and strong selection baselines, achieving up to $+2.45$ average accuracy improvement over the strongest baseline with the same rollout budget.
\end{itemize}

\begin{figure}[t]
    \centering
    \includegraphics[width=0.99\linewidth]{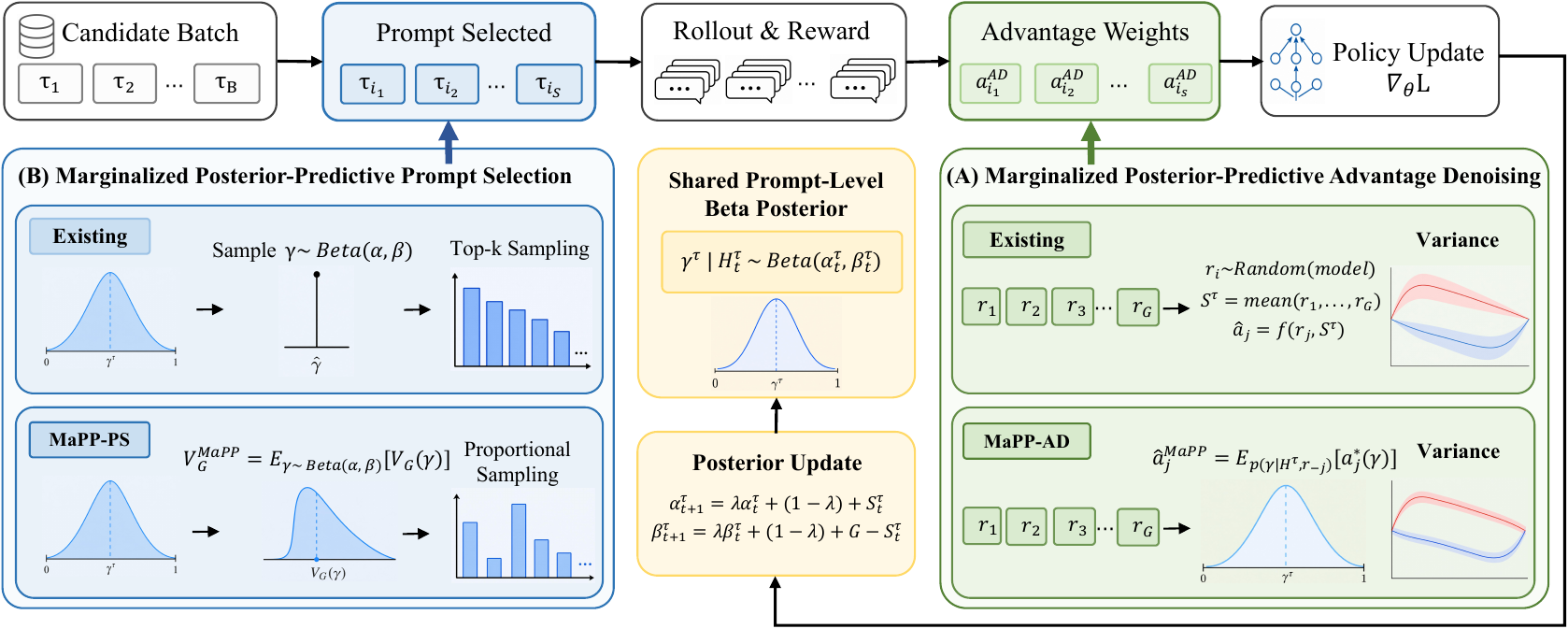}
    \caption{\textbf{Overview of the MaPP framework.} A shared Beta posterior $(\alpha_t^{\tau}, \beta_t^{\tau})$, updated online from rollout history, serves as the foundation for two closed-form components. \textbf{(A) MaPP-AD} marginalizes $a_j^{\star}(\gamma)$ over a leave-one-out posterior, replacing the composition-dependent GRPO advantage with composition-invariant weights. Both components reuse the same posterior at no additional rollout cost. \textbf{(B) MaPP-PS} marginalizes $V_G(\gamma)$ over the full posterior for uncertainty-aware prompt selection, replacing the point-estimate scoring used by existing methods.}
    \label{fig:main}
    \vspace{-20pt}
\end{figure}

\section{Related Work}
\label{sec:related_work}

\subsection{RL Finetuning for Large Language Models}
\label{sec:rw_rl}


RLVR is a dominant post-training paradigm for reasoning LLMs~\citep{deepseek,deepseekmath,jaech2024openai}. GRPO~\citep{deepseekmath} eliminates PPO's~\citep{ppo} learned value network via within-group normalization, setting the template for many follow-ups.  
Variants address specific failure modes: Dr.GRPO~\citep{liu2025understanding} removes length normalization and std scaling, DAPO~\citep{dapo} adds asymmetric clipping and dynamic sampling, VAPO~\citep{vapo} reintroduces a value model for long CoT. Others include sequence-level importance ratios~\citep{zheng2025group} and adaptive reweighting~\citep{yang2026your}.  
These modify \emph{what} is optimized. MaPP is complementary: it modifies \emph{how} the GRPO advantage is estimated under group composition uncertainty, while operating within the same GRPO training framework. Conceptually, MaPP Rao--Blackwellizes~\citep{casella1996rao} the GRPO advantage using a conjugate Beta posterior instead of a learned critic~\citep{liu2017action}.

\subsection{Online Prompt Selection for RLVR}
\label{sec:rw_selection}

Prompts in RLVR contribute unevenly: overly easy or extremely hard prompts yield vanishing gradients, while intermediate ones provide informative signals~\citep{dapo,bae2026online,zeng2025cures}. This motivates online selection methods~\citep{dapo,mopps,dps,greso,insight,gps}.

Existing approaches fall into two paradigms~\citep{gps}: evaluation-based (e.g., Dynamic Sampling~\citep{dapo}, SPEED-RL~\citep{zhang2025speed}) oversample and filter, wasting budget; prediction-based (MoPPS~\citep{mopps}, DPS~\citep{dps}, GRESO~\citep{greso}, GPS~\citep{gps}, INSIGHT~\citep{insight}) estimate difficulty before rollout. These mainly improve budget allocation across prompts, but ignore composition noise from uncertainty in group composition when GRPO assigns response-level advantages.

\xu{Unlike these methods, MaPP also addresses signal extraction from sampled responses. Specifically, {MaPP-AD} denoises the GRPO advantage via posterior-predictive marginalization, and {MaPP-PS} reuses the shared Beta posterior for a more faithful prompt-selection score. Both are modular and can be layered on existing pipelines.}

\section{Challenge Analysis}
\label{Observation}
\subsection{Preliminaries}

\noindent\textbf{GRPO for RLVR: } Given a prompt $\tau$ from training dataset $\mathcal{T} = \{\tau_i\}_{i=1}^{N}$,where $N$ is the size of the training dataset, and $G$ independent 
responses $\{y_j\}_{j=1}^G$ generated by the policy 
$\pi_\theta(\cdot \mid \tau)$, each receiving a binary 
reward $r_j^{\tau} \in \{0,1\}$ verifying 
correctness~\citep{deepseekmath,deepseek}, the objective of RL finetuning is to maximize the expected return:
\begin{equation}
\max_{\theta}\;
\mathbb{E}_{\tau \sim \mathcal{T},\; \{y_j\}_{j=1}^G \sim \pi_\theta(\cdot \mid \tau)}
\left[\frac{1}{G}\sum_{j=1}^G r_j^\tau\right],
\label{eq:rl_obj}\end{equation}
where $G$ denotes the group size.
GRPO~\citep{deepseekmath} eliminates the need for a learned value network by estimating advantages through within-group normalization.
For each prompt $\tau$, the policy generates a group of $G$ independent responses $\{y_j\}_{j=1}^G$, each receiving a binary reward $r_j^{\tau} \in \{0,1\}$. The group-relative advantage for the $j$-th response is then defined as:
\begin{equation}
\hat{A}_j^{\tau} = \frac{r_j^{\tau} - \mathrm{mean}(\{r_k^{\tau}\}_{k=1}^{G})} {\mathrm{std}(\{r_k^{\tau}\}_{k=1}^{G})}.
\label{eq:grpo_advantage}
\end{equation}
As noted in DAPO~\citep{dapo}, when all responses to a given prompt are either all correct or all incorrect, they receive identical rewards, yielding $\hat{A}_j^{\tau} = 0$. In this scenario, the prompt is \emph{degenerate} and contributes no gradient signal~\citep{dapo,greso}. To identify such conditions, prior work on prompt selection~\citep{greso} introduces a success count variable $S^{\tau} \in \{0, G\}$ indicating the degenerate case, \xu{{\em i.e.}, all responses are correct or all are incorrect.}

\noindent\textbf{Online Prompt Selection: } Prediction-based online prompt selection introduces a success rate $\gamma^{\tau} \in [0,1]$ for each prompt $\tau$ to estimate prompt difficulty for selection.
Since $\gamma$ is not directly observable, prior works~\citep{mopps,insight} first model each response reward $r_j^{\tau} \in \{0,1\}$ generated from prompt $\tau$ as an independent Bernoulli trial:
\begin{equation}
r_j^{\tau} \mid \gamma^{\tau} \sim \mathrm{Bernoulli}(\gamma^{\tau}) \quad \forall j \in \{1, \cdots G\}.
\label{eq:bernoulli}
\end{equation}
\xu{We define the corresponding success count as} 
$S^{\tau} := \sum_{j=1}^{G} r_j^{\tau}$, 
which follows a Binomial distribution, {\em i.e.}, 
$S^{\tau} \mid \gamma^{\tau} \sim \mathrm{Binomial}(G,\, \gamma^{\tau})$.

Then, a recursive Bayesian update procedure is applied for efficient posterior inference of the success rates $\gamma^{\tau}$. At the initial stage a Beta prior models the initial success rate formulated as:
\begin{equation}
    \gamma^{\tau}_0 \sim \mathrm{Beta}(\alpha^{\tau}_0, \beta^{\tau}_0),
\end{equation}
where $\alpha^{\tau}_0$ and $\beta^{\tau}_0$ reflect prior pseudo-counts of successes and failures, typically set to $(1, 1)$ for a uniform prior.
During prompt selection, the optimization history is denoted as $\mathcal{H}_{t}= \{B_t, R_t\}$, where $B_t$ is the selected prompt batch at step $t$ and $R_t = \{r_j^{\tau}\}_{\tau \in B_t}^{j \in G}$ collects the corresponding rewards.
By Bayes rule, the posterior distribution over $\gamma^{\tau}_t$ given observations up to step $t$ is:
\begin{equation}
    p(\gamma^{\tau}_t \mid \mathcal{H}_t) \propto p(r_j^{\tau} \mid \gamma^{\tau}_t) \cdot p(\gamma^{\tau}_t \mid \mathcal{H}_{t-1}),
    \label{eq:beta_posterior}
\end{equation}
where $p(\gamma^{\tau}_t \mid \mathcal{H}_{t-1}) \sim \mathrm{Beta}(\alpha^{\tau}_t, \beta^{\tau}_t)$ represents the conditional prior using the last updated posterior as the proxy. $p(r_j^{\tau} \mid \gamma^{\tau}_t)$ is the likelihood of observing the feedback under prompt $\tau$ at step $t$.

With a temporal discounting factor $\lambda \in (0,1)$ applied for tracking the evolving policy~\citep{mopps,insight}, the posterior parameters are updated as:
\begin{equation}
\alpha^{\tau}_{t+1} = \lambda \cdot \alpha^{\tau}_t + (1{-}\lambda) \cdot \alpha^{\tau}_0 + S^{\tau}_t, \qquad \beta^{\tau}_{t+1} = \lambda \cdot \beta^{\tau}_t + (1{-}\lambda) \cdot \beta^{\tau}_0 + G - S^{\tau}_t.
\label{eq:beta_update}
\end{equation}
Existing selection methods use this posterior to prioritize prompts with intermediate difficulty, as these are most likely to yield non-degenerate gradients~\citep{bae2026online,mopps,gps}.

\subsection{Composition Noise in GRPO}
\label{sec:motivation}
Previous online prompt selection methods model prompt difficulty using success rates and update it via posterior probabilities~\citep{mopps,dps,greso,insight,gps}, while suffer from {\em composition noise} as illustrated in Fig.~\ref{fig:teaser} (b). For an in-depth analysis, we formulate the explicit form of the aforementioned {\em composition noise}. The variance of the advantage for each response under the same prompt $\tau$ admits a precise decomposition via the law of total variance, expressed as:
\begin{equation}
\underbrace{\mathrm{Var}\!\big(\hat{A}^{\tau}_j \mid \gamma^{\tau} \big)}_{\text{total}} \;=\; \underbrace{\mathrm{Var}\!\big(\mathbb{E}[\hat{A}^{\tau}_j \mid r^{\tau}_j, \gamma^{\tau}] \,\big|\, \gamma^{\tau} \big)}_{\text{composition-free signal}} \;+\; \underbrace{\mathbb{E}\!\big[\mathrm{Var}(\hat{A}^{\tau}_j \mid r^{\tau}_j, \gamma^{\tau}) \,\big|\, \gamma^{\tau} \big]}_{\text{composition noise}\;\xi(\gamma^{\tau})}.
\label{eq:eve}
\end{equation}
The first term describes the variation in advantage among different responses to the same prompt, which distinguishes correct responses from incorrect ones, reflecting the genuine signal.
The second term describes that, even when the response is fixed as correct or incorrect, its advantage still varies across different groups — a variation that arises entirely from the random composition of the group. This forms the {\em composition noise} term $\xi(\gamma^{\tau})$, which is intrinsic and independent to prompt selection.
However, $\xi(\gamma^{\tau})$ is strictly positive for all $\gamma^{\tau} \in (0,1)$ (Theorem~\ref{thm:floor}) and cannot be reduced by prompt selection alone. 
%
\xu{In standard GRPO, the realized advantage $\hat{A}_j^{\tau}$ directly enters the policy gradient, so the composition-noise term in Eq.\eqref{eq:eve} contributes variance to gradient estimation. Since prompt-selection methods assess prompt informativeness through sampled group rewards generated by the same policy, this noise also makes such assessments less faithful.}
\xu{Therefore, rather than using the noisy realized advantage $\hat{A}_j^{\tau}$, we target its conditional expectation $\mathbb{E}[\hat{A}_j^{\tau} \mid r_j^{\tau}, \gamma^\tau]$, which depends only on the response reward and the prompt’s latent difficulty and removes the composition-noise term by construction.}

\section{Method}
\label{sec:method}


The decomposition in Eq.~\eqref{eq:eve} identifies $\mathbb{E}[\hat{A}^\tau_j \mid r^\tau_j, \gamma^\tau]$ as the composition-free target. This section formalizes that target (\S\ref{sec:intrinsic}), recovers it via posterior-predictive marginalization (\S\ref{sec:recovery}), and extends the same principle to prompt selection (\S\ref{sec:pp_selection}). \xu{In this section, we discuss our method with a given prompt $\tau$. Hence, we omit superscript $\tau$ for simplicity.}

\subsection{The Intrinsic Advantage}
\label{sec:intrinsic}


The intrinsic advantage answers the question the standard GRPO advantage fails to: \emph{given that this response is correct (or not) on a prompt of difficulty $\gamma$, what advantage should it receive on average?} Intuitively, it should depend only on the response reward and the prompt intrinsic difficulty, never on which peers happened to be drawn. Therefore, we first give the definition of the intrinsic advantage $a_j^{\star}(\gamma)$ and its closed form below. 

\begin{definition}[Intrinsic advantage]
\label{def:intrinsic}
For a prompt $\tau$ with success rate $\gamma \in (0,1)$, the \emph{intrinsic advantage} of response $j$ is
\begin{equation}
a_j^{\star}(\gamma^{}) \;:=\; \mathbb{E}\!\big[\hat{A}_j \,\big|\, r_j^{},\, \gamma^{}\big],
\label{eq:intrinsic_def}
\end{equation}

\end{definition}

\begin{proposition}[Closed form]
\label{prop:intrinsic}
Under $r_j^{} \mid \gamma^{} \stackrel{\mathrm{i.i.d.}}{\sim} \mathrm{Bernoulli}(\gamma^{})$,
\begin{equation}
a_j^{\star}(\gamma^{}) = \begin{cases} +\mu_G(\gamma^{}), & r_j^{} = 1,\\[4pt] -\mu_G(1 - \gamma^{}), & r_j^{} = 0,
\end{cases}
\label{eq:intrinsic_closed}
\end{equation}
where 
\begin{equation}
\mu_G(p) \;=\; \sum_{S^{}=1}^{G-1}\binom{G-1}{S^{}-1}\,p^{S^{}-1}(1{-}p)^{G-S^{}}\, \sqrt{\frac{G{-}S^{}}{S^{}}},\,\,G\geq 2.
\label{eq:mu_G}
\end{equation}
\end{proposition}
\xu{Note that in the degenerate cases where $S^{}=\{0,G\}$ contribute zero and are omitted from the sum.}
Applying such intrinsic advantage $a_j^{\star}(\gamma^{})$ for $\hat{A}_j$ guarantees that $\operatorname{Var}[a_j^{\star} \mid r_j, \gamma^{}] = 0$. Therefore, the total variance in Eq.\eqref{eq:eve} degenerates into the composition-free signal term alone and intrinsically eliminates the composition-noise term. 
However, $a_j^{\star}$ depends on the latent success rate $\gamma^{}$, which can not be observed or explicitly derived. Therefore, we recover it through closed-form marginalization over the Beta posterior.

\subsection{Marginalized Posterior-Predictive Advantage Denoising}
\label{sec:recovery}



To avoid each response's self-influence~\citep{liu2025understanding}, we marginalize the latent success rate $\gamma^{}$ over a leave-one-out Beta posterior that excludes response $j$'s own reward, formulated as:
\begin{equation}
p(\gamma_{} \mid \mathcal{H}_t, r_{t;-j}) \;=\; \mathrm{Beta}(\alpha_{t+1;j}',\, \beta_{t+1;j}'), 
\end{equation}
\begin{equation}
\alpha_{t+1;j}' := \alpha_{t+1} {-} r_{t;j},\;\; \beta_{t+1;j}' := \beta_{t+1} - 1 + r_{t;j},
\label{eq:loo_posterior}
\end{equation}
\xu{where group-shared $\alpha_{t+1}$ and $\beta_{t+1}$ are calculated according to Eq.~(\ref{eq:beta_update}).}
\xu{$r_{t;-j}$ denotes the other rewards in the group except for $j$-th response at $t$-th iteration.} 
%
%
%
This construction guarantees that the posterior used to estimate the advantage of response $j$ is conditionally independent of $r_j$; consequently, the resulting estimator is free from bias induced by the response it evaluates.
%
At each training step $t$, we marginalize $a_j^{\star}(\gamma)$ over the leave-one-out posterior to obtain a composition-free advantage estimate:
\begin{equation}
\hat{a}_{j}^{\mathrm{MaPP}} \;:=\; 
\mathbb{E}_{p(\gamma \mid \mathcal{H}_t, 
r_{t;-j})}\!\big[a_j^{\star}(\gamma)\big] \;=\; 
\int_0^1 a_{j}^{\star}(\gamma) \; 
p(\gamma \mid \mathcal{H}_t, r_{t;-j}) 
\;\mathrm{d}\gamma.
\label{eq:pp_def}
\end{equation}
With Eq.\eqref{eq:pp_def}, we average all plausible values of $\gamma^{}$, which is weighted by the observed optimization history and forms $\hat{a}_j^{\mathrm{MaPP}}$, called MaPP-AD. Consequently, the closed form of our tractable integral MaPP-AD is given by Beta--Binomial conjugacy. 
\begin{proposition}[Closed form]
\label{prop:pp_closed}
At training step $t$, let $\alpha_j' := \alpha_{t+1;j}'$ 
and $\beta_j' := \beta_{t+1;j}'$ for brevity. Then
\begin{equation}
\hat{a}_j^{\mathrm{MaPP}} = \begin{cases} 
+\displaystyle\sum_{S=1}^{G-1} 
w_{S}(\alpha_j',\, \beta_j') 
\,\sqrt{\dfrac{G{-}S}{S}}, & r_j = 1,\\[14pt] 
-\displaystyle\sum_{S=1}^{G-1} 
w_{S}(\beta_j',\, \alpha_j') 
\,\sqrt{\dfrac{G{-}S}{S}}, & r_j = 0,
\end{cases}
\label{eq:pp_closed}
\end{equation}
where $w_S(\alpha, \beta) := P_{\mathrm{BB}}(S - 1 \mid G - 1,\, \alpha, \beta) = \binom{G-1}{S-1} \dfrac{B(\alpha + S - 1,\, \beta + G - S)}{B(\alpha, \beta)}$ is the Beta-Binomial PMF evaluated at $S - 1$ successes out of $G - 1$ trials.
\end{proposition}

The weight $w_S(\alpha_j', \beta_j')$ has a concrete meaning: it is the posterior-predictive probability that a hypothetical fresh group of $G{-}1$ responses would contain $s{-}1$ correct ones. The estimator evaluates the advantage at every plausible composition $s$ and averages under the posterior belief, replacing the single realized composition with its full predictive distribution. As the posterior concentrates around the true pass rate with more observations, $\hat{a}_j^{\mathrm{MaPP}}$ converges to the intrinsic advantage $a_j^{\star}(\gamma)$. 
We show in Appendix~\ref{app:proofs} that $\hat{a}_j^{\mathrm{MaPP}}$ achieves provably lower MSE than GRPO once the posterior accumulates a few updates per prompt.

\subsection{Marginalized Posterior-Predictive Prompt Selection}
\label{sec:pp_selection}

\S\ref{sec:intrinsic}--\ref{sec:recovery} \xu{focused on denoising response-level advantage estimation. We now show how the same posterior-predictive principle also yields a more faithful prompt-selection score.}
Existing selection methods score prompts via point estimates of difficulty and select the top-$k$ candidates closest to a target difficulty~\citep{mopps,dps,greso,insight}. This point-estimate approach mirrors the same limitation exposed in \S\ref{sec:motivation}: a single value of $\gamma$ is substituted for the latent quantity, discarding posterior uncertainty.

We propose instead a principled informativeness score derived from the same marginalization principle. Define the probability that a prompt yields a non-degenerate gradient as $V_G(\gamma) = 1 - (\gamma)^G - (1-\gamma)^G$, which peaks at $\gamma = 1/2$ and vanishes at the extremes. Since $\gamma$ is latent, we marginalize against the posterior:
\begin{equation}
V_G^{\mathrm{MaPP}}(\alpha^{\tau}, \beta^{\tau}) \;:=\; \mathbb{E}_{\gamma \sim p(\gamma \mid \mathcal{H}^{\tau})}\!\big[V_G(\gamma)\big] \;=\; 1 - \frac{B(\alpha^{\tau}{+}G,\, \beta^{\tau})}{B(\alpha^{\tau}, \beta^{\tau})} - \frac{B(\alpha^{\tau},\, \beta^{\tau}{+}G)}{B(\alpha^{\tau}, \beta^{\tau})}.
\label{eq:vg_pp}
\end{equation}
Since $V_G$ is concave, Jensen's inequality gives $V_G^{\mathrm{MaPP}} \leq V_G(\mathbb{E}[\gamma^{\tau} \mid \mathcal{H}^{\tau}])$: the marginalized score automatically down-weights prompts whose posterior is wide, avoiding wasted rollouts on prompts that appear informative under a point estimate but whose true difficulty remains uncertain. 
Batches are drawn by sampling prompts with probability proportional to $V_G^{\mathrm{MaPP}}(\alpha^{\tau}, \beta^{\tau})$, preserving exploration over prompts whose posterior is still consolidating.

\paragraph{Unified posterior-predictive gradient.} Combining posterior-predictive selection with the posterior-predictive advantage of \S\ref{sec:recovery}, the MaPP gradient takes the form
\begin{equation}
\nabla J
=
\mathbb{E}_{\tau \sim q,\; \{y_j^\tau\}_{j=1}^G \sim \pi_\theta(\cdot \mid \tau)}
\left[
\frac{1}{G}\sum_{j=1}^{G}
\hat a_j^{\mathrm{MaPP},\tau}\,
\nabla_\theta \log \pi_\theta(y_j^\tau \mid \tau)
\right],
\qquad
q(\tau)\propto V_G^{\mathrm{MaPP}}(\alpha^\tau,\beta^\tau).
\label{eq:final_grad}
\end{equation}
\xu{where both the batch distribution $q(\tau)$ and the per-response weight $a_j^{\mathrm{MaPP}}$ are posterior-predictive marginalizations against the same shared Beta posterior.}
Both components are modular: each can be adopted independently or composed with existing selection strategies, adding $O(|\mathcal{B}| \cdot G^2)$ per-step overhead, negligible compared to rollout and backpropagation. Algorithm~\ref{alg:method} gives the full training loop.

\section{Experiments}

\begin{table*}[t]
\caption{Evaluation results on CountDown3to4 and mathematical reasoning benchmarks. \textbf{Bold} indicates the best result. CountDown is evaluated after CountDown3to4 training, while all other benchmarks use math-trained models. Rollouts and GPU hours report math-training cost.}
\label{tab:math}
\centering
\resizebox{\textwidth}{!}{%
\small
\setlength{\tabcolsep}{4pt}
\begin{tabular}{ll ccccc c cc}
\toprule
{Models} & {Methods}
  & {CountDown} & {AMC} & {MATH500} & {Minerva.} & {Olympiad.}
  & {Avg.$\uparrow$} & {Rollouts$\downarrow$} & {GPU Hours$\downarrow$} \\
\midrule
\multirow{5}{*}{{Qwen3-4B}}
 & {Random}  & 80.95 & 54.22 & 78.63 & 28.31 & 38.25 & 56.07 & 563k & 90h \\
 & {MoPPS}   & 83.68 & 60.24 & 81.04 & 28.31 & 44.28 & 59.51 & 563k & 81h \\
 & {DPS}     & 83.42 & 59.04 & 80.65 & 28.68 & 43.98 & 59.15 & 563k & 81h \\
 & {{DS}}
   & {83.83} & {60.24} & {82.06} & {30.88} & {45.03} & {60.41} & {2252k} & {209h} \\
 & {{\textbf{Ours}}}
   & \textbf{85.67}
   & \textbf{62.65}
   & \textbf{82.66}
   & \textbf{31.25}
   & \textbf{47.59}
   & \textbf{61.96}
   & \textbf{563k} & \textbf{81h} \\
\midrule
\multirow{5}{*}{{Qwen3-8B}}
 & {Random}  & 81.29 & 55.42 & 81.05 & 30.51 & 42.17 & 58.09 & 563k & 104h \\
 & {MoPPS}   & 84.69 & 62.65 & 82.46 & 30.15 & 50.60 & 62.11 & 563k & 88h \\
 & {DPS}     & 85.80 & 60.24 & 81.45 & 29.78 & 48.49 & 61.15 & 563k & 89h \\
 & {{DS}}
   & {83.79} & {61.45} & {82.86} & {\textbf{31.62}} & {48.34} & {61.61} & {2252k} & {230h} \\
 & {{\textbf{Ours}}}
   & \textbf{87.51}
   & \textbf{63.86}
   & \textbf{83.47}
   & 30.88
   & \textbf{52.26}
   & \textbf{63.60}
   & \textbf{563k} & \textbf{88h} \\
\midrule
\multirow{5}{*}{{R1-Distill-7B}}
 & {Random}  & 78.92 & 57.83 & 80.24 & 27.57 & 38.86 & 56.68 & 563k & 63h \\
 & {MoPPS}   & 80.75 & 59.20 & 81.45 & 26.84 & 41.11 & 57.87 & 563k & 59h \\
 & {DPS}     & 80.03 & 59.04 & 79.64 & 25.74 & 40.66 & 57.02 & 563k & 58h \\
 & {{DS}}
   & {81.52} & {\textbf{62.65}} & {80.85} & {27.81} & {41.57} & {58.88} & {2231k} & {152h} \\
 & {{\textbf{Ours}}}
   & \textbf{82.24}
   & 61.45
   & \textbf{82.26}
   & \textbf{28.31}
   & \textbf{42.17}
   & \textbf{59.29}
   & \textbf{563k} & \textbf{58h} \\
\bottomrule
\end{tabular}
}
\vspace{-5pt}
\end{table*}

\subsection{Experimental Setup}
\label{sec:setup}

We evaluate MaPP across three reasoning domains, mathematics, numerical planning, and visual geometry, using diverse model backbones to assess generality.

\paragraph{Tasks.}

For \textbf{Mathematics}, we train on MATH~\citep{MATH} (7,500 problems) and evaluate on AMC23, MATH500~\citep{MATH500}, Minerva Math~\citep{Minerva}, and OlympiadBench~\citep{Olym}.
For \textbf{Numerical planning}, we use the Countdown Number Game: train on Countdown-34~\citep{Countdown}, evaluate on held-out CD-34 and harder CD-4 (four source numbers, larger search space).
For \textbf{Visual geometry}, we train and evaluate on Geometry3k~\citep{lu2021intergps,hiyouga2025geo3k}, pairing diagrams with multi-step reasoning questions.

\paragraph{Models.}
For mathematics and planning, we use three text-only models: Qwen3-4B-Base, Qwen3-8B-Base~\citep{yang2025qwen3}, and DeepSeek-R1-Distill-Qwen-7B~\citep{deepseek}. 
For visual geometry, we adopt Qwen2.5-VL-3B-Instruct and Qwen2.5-VL-7B-Instruct~\citep{bai2025qwen25vl}.

\paragraph{Baselines.}
We compare against four methods spanning online prompt selection paradigms~\citep{gps}.
\textbf{(1) Random}: uniform sampling (standard GRPO baseline)~\citep{deepseekmath}.
\textbf{(2) MoPPS}~\citep{mopps}: prediction-based; models success rate as Beta variable, uses Thompson sampling for intermediate difficulty.
\textbf{(3) DPS}~\citep{dps}: prediction-based; frames solving progress as a 3-state HMM, prioritizes partially-solved prompts.
\textbf{(4) Dynamic Sampling (DS)}~\citep{dapo}: evaluation-based; oversamples $4\times$ batch, rolls out all, filters zero-variance prompts. Serves as compute-intensive oracle baseline.


\begin{figure*}[t]
    \centering
    \includegraphics[width=\textwidth]{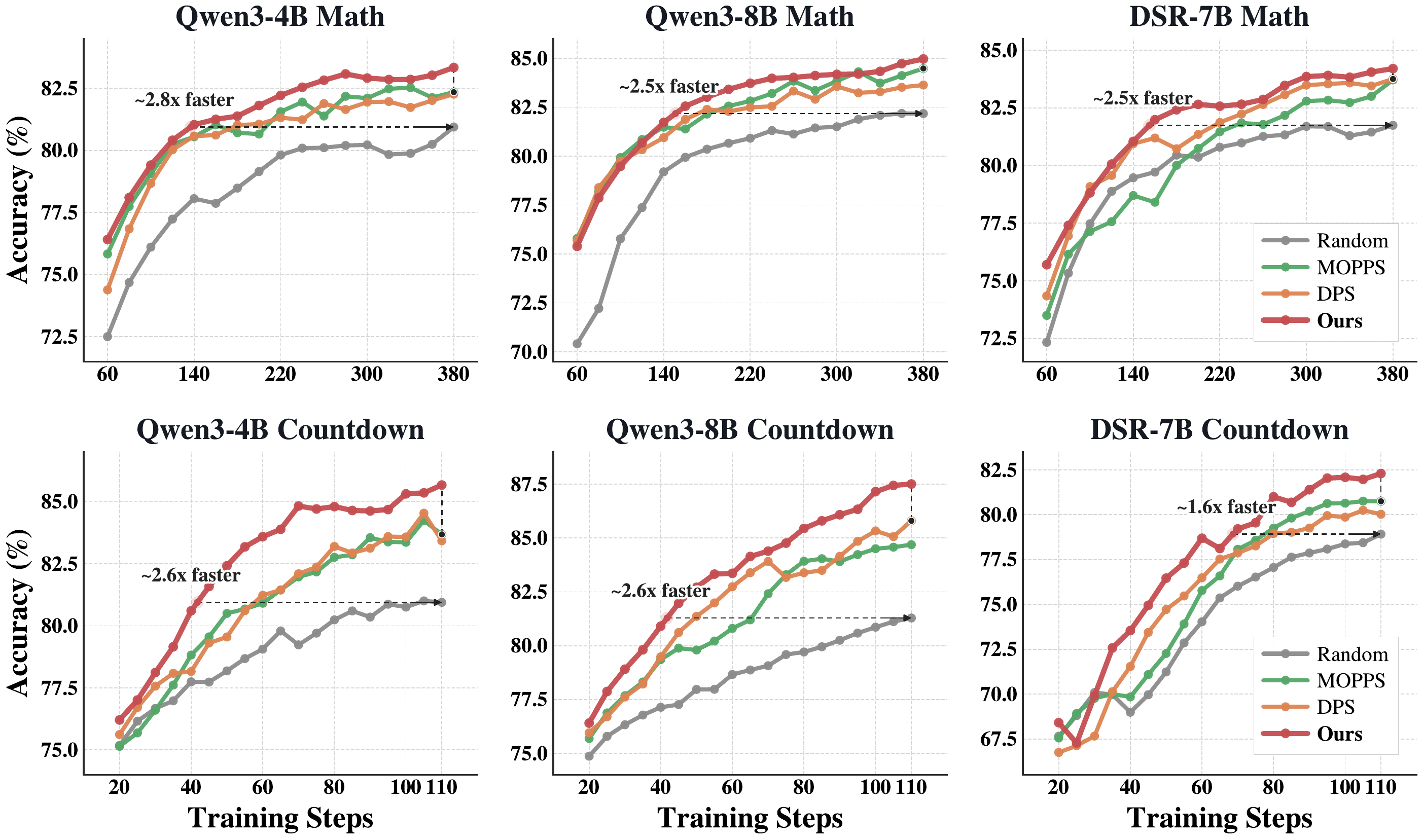}
    \caption{Performance comparisons of different methods and models on the MATH and Countdown tasks. Our proposed MaPP outperforms the existing SOTA methods and other baselines in both training efficiency and performance.}
    \label{fig:main_exp}
    \vspace{-10pt}
    
\end{figure*}

\subsection{Main Results}
\paragraph{Mathematics.} Tab.~\ref{tab:math} reports evaluation results across five mathematics benchmarks for three model backbones. Across all scales, MaPP consistently achieves the highest average accuracy, outperforming both prediction-based methods (MoPPS, DPS) and the compute-intensive oracle baseline DS. On Qwen3-4B, MaPP obtains an average score of 61.96, improving over the strongest prediction-based baseline MoPPS by {+2.45} and even surpassing DS by {+1.55}, while using only 25\% of its rollout budget (563k vs.\ 2252k). Similar trends hold on Qwen3-8B ({+1.49} over MoPPS) and R1-Distill-7B ({+1.42} over MoPPS), confirming that the improvement is consistent across model families and scales.

\paragraph{Planning and visual geometry.} On the Countdown task (Fig.~\ref{fig:main_exp}, bottom row), MaPP achieves final accuracy gains of {+1.84} (Qwen3-4B) and {+1.71} (Qwen3-8B) over the strongest baseline at convergence. For visual geometry on Geometry3k (Fig.~\ref{fig:abtion_lambda}(a)), MaPP outperforms the strongest baseline by {+1.04} on Qwen2.5-VL-3B-Instruct and {+2.40} on Qwen2.5-VL-7B-Instruct. These results demonstrate that MaPP generalizes beyond text-only mathematical reasoning to both planning and multi-modal tasks.

\paragraph{Training efficiency.} Fig.~\ref{fig:main_exp} plots accuracy against training steps across all six model-task combinations. MaPP converges substantially faster than all baselines: on MATH, {it reaches the same accuracy as Random Sample by approximately {2.5--2.8$\times$} faster convergence speed}, {while the acceleration rates range from {1.6$\times$} to {2.6$\times$} on Countdown dataset.} MaPP introduces no additional rollout cost over prediction-based methods, and requires less than half the runtime of DS across all settings (Tab.~\ref{tab:math}, last two columns).

\paragraph{Discussion.} Several patterns emerge from the results. 
First, the gains are most pronounced on challenging benchmarks such as OlympiadBench (+3.31 on Qwen3-4B), suggesting that cleaner gradient signals from advantage denoising disproportionately benefit the learning of harder reasoning skills.
Second, MaPP {significantly} improves over MoPPS and DPS despite sharing the same Beta posterior infrastructure, confirming that the gains stem from how the posterior is used, specifically marginalization for both advantage estimation and prompt selection, rather than from a better posterior itself.
Third, the improvement persists across three task families and five model backbones, indicating that posterior-predictive marginalization provides a general refinement to GRPO learning signals rather than a task-specific heuristic.

\begin{table}[t]
\centering
\footnotesize
\begin{minipage}[t]{0.46\columnwidth}
\centering
\caption{Sensitivity to the temporal discount $\lambda$. Trained on Qwen3-4B, Countdown.}
\label{tab:ablation_lambda}
\setlength{\tabcolsep}{6pt}
\renewcommand{\arraystretch}{1.05}
\begin{tabular}{c cc c}
\toprule
$\lambda$ & {CountDown3to4} & {CountDown4} & {Avg.} \\
\midrule
0.0   & 83.51 & 66.19 & 74.85 \\
0.1   & 85.17 & 70.04 & 77.61 \\
\textbf{0.5}   & \textbf{86.76} & \textbf{72.24} & \textbf{79.50} \\
0.7   & 85.95 & 69.18 & 77.57 \\
0.9   & 83.08 & 67.33 & 75.21 \\
1.0   & 85.82 & 70.72 & 78.27 \\
\bottomrule
\end{tabular}
\end{minipage}
\hfill
\begin{minipage}[t]{0.52\columnwidth}
\centering
\caption{Robustness to rollout group size $G$.
Trained on Qwen3-4B, Countdown.}
\label{tab:ablation_G}
\setlength{\tabcolsep}{2.4pt}
\begin{tabular}{l ccc ccc}
\toprule
& \multicolumn{3}{c}{$G=4$} & \multicolumn{3}{c}{$G=16$} \\
\cmidrule(lr){2-4} \cmidrule(lr){5-7}
{Method} & {CD-34} & {CD-4} & {\#Rollouts} & {CD-34} & {CD-4} & {\#Rollouts} \\
\midrule
Random       & 79.39 & 60.14 & 123k & 82.54 & 64.01 & 492k \\
MoPPS        & 80.22 & 62.70 & 123k & 86.39 & 72.79 & 492k \\
DPS          & 82.17 & 63.26 & 123k & 86.90 & 70.18 & 492k \\
DS  & 81.85 & 64.11 & 490k & 83.87 & 66.66 & 1855k \\
Ours         & \textbf{82.45} & \textbf{64.44} & 123k & \textbf{88.33} & \textbf{73.42} & 492k \\
\bottomrule
\end{tabular}
\end{minipage}
\vspace{-10pt}
\end{table}
\subsection{Ablation Analysis}
\label{sec:ablation}

\begin{wraptable}{r}{0.46\textwidth}
\vspace{-12pt}
\caption{\textbf{Ablation on the two components.} Trained on MATH, Qwen3-4B.
MaPP-AD: posterior-predictive advantage (§~\ref{sec:recovery}). 
MaPP-PS: posterior-predictive selection (§~\ref{sec:pp_selection}).}
\label{tab:ablation_2x4}
\centering
\small
\setlength{\tabcolsep}{6pt}
\begin{tabular}{l cc c}
\toprule
{Method} & {MATH} & {Olympiad.} & {Avg.} \\
\midrule
{GRPO}                          & 79.41 & 38.25 & 58.83 \\
~~+ MaPP-PS                    & 81.03 & 43.17 & 62.10 \\
~~+ MaPP-AD                                & 82.07 & 44.52 & 63.30 \\
~~\textbf{+ AD + PS}             & \textbf{82.85} & \textbf{47.59} & \textbf{65.22} \\
\cmidrule(lr){1-4}
\noalign{\vskip 3pt}
{MoPPS}                         & 82.17 & 44.28 & 63.23 \\
~~+ MaPP-AD                                & \textbf{82.63} & \textbf{45.90} & \textbf{64.27} \\
\cmidrule(lr){1-4}
\noalign{\vskip 3pt}
{DPS}                           & 82.24 & 43.98 & 63.11 \\
~~+ MaPP-AD                                & \textbf{82.71} & \textbf{45.75} & \textbf{64.23} \\
\bottomrule
\end{tabular}
\vspace{-15pt}
\end{wraptable}
\paragraph{Effect of the two components.} MaPP comprises two modules: the posterior-predictive advantage (MaPP-AD) and the posterior-predictive selection score (MaPP-PS). 
Tab.~\ref{tab:ablation_2x4} disentangles the contributions of MaPP-AD and MaPP-PS on Qwen3-4B. Starting from standard GRPO, adding MaPP-PS alone yields +1.62 improvement on MATH, while MaPP-AD alone yields a larger +2.66. Combining both produces +3.44, exceeding the sum of individual gains, suggesting the two modules are synergistic. 
To verify complementarity with existing methods, we plug MaPP-AD into MoPPS and DPS without modifying their selection logic; it consistently improves both (+1.04 and +1.12 respectively), demonstrating that advantage denoising provides additional gains beyond prompt selection.

\paragraph{Sensitivity to the temporal discount $\lambda$.} The temporal discount $\lambda$ controls the trade-off between tracking speed and posterior stability. Tab.~\ref{tab:ablation_lambda} and Fig.~\ref{fig:abtion_lambda}(b) reports MaPP performance on Countdown under varying $\lambda$. Performance peaks at $\lambda{=}0.5$ (79.50 avg.) and degrades at both extremes: $\lambda{=}0.0$ (74.85) forgets all history and reduces the posterior to single-step evidence, while $\lambda{=}0.9$ (75.21) retains stale observations that lag behind the evolving policy. The moderate sensitivity suggests that a reasonable balance between recency and stability is sufficient, and the optimal value does not require fine-grained tuning. We use $\lambda{=}0.5$ for all other experiments.

\paragraph{Robustness to group size.} Tab.~\ref{tab:ablation_G} and Fig.~\ref{fig:abtion_group} evaluate MaPP on the Countdown benchmark under group sizes $G{=}4$ and $G{=}16$, complementing the default $G{=}8$ used on in main experiment. MaPP achieves the highest accuracy across both settings: at $G{=}4$, it outperforms the best baseline by +0.28 (CD-34) and +0.33 (CD-4); at $G{=}16$, the margin widens to +1.43 (CD-34) and +0.63 (CD-4).
The wider margin at larger $G$ is expected: more responses per group provide richer per-step evidence for the posterior, enabling more precise advantage corrections that outweigh the averaging effect of larger groups.
This confirms that posterior-predictive marginalization scales favorably with group size.

\section{Conclusion}
\label{sec:conclusion}


We identified composition noise in the GRPO advantage due to uncertainty in group composition, which degrades gradient estimation and prompt selection. To address this, we proposed MaPP: a unified framework that extends the per-prompt Beta posterior to a closed-form posterior-predictive advantage estimator with provably diminishing MSE and an uncertainty-aware selection score, addressing both data-efficiency losses through a single shared posterior. Experiments on mathematics, planning, and visual geometry with five backbones show consistent gains in accuracy and convergence over GRPO and strong baselines, with no additional rollout overhead.



{\small
\bibliographystyle{abbrvnat}
\bibliography{reference}
}

\newpage

\newpage
\renewcommand\thefigure{\Alph{figure}} 
\setcounter{figure}{0}
\renewcommand\thetable{\Alph{table}} 
\setcounter{table}{0}
\renewcommand{\thesection}{\Alph{section}}

\setcounter{section}{0}
\setcounter{theorem}{0}

\renewcommand{\theHsection}{\Alph{section}}
\renewcommand{\theHsubsection}{\Alph{section}.\arabic{subsection}}
\renewcommand{\theHsubsubsection}{\Alph{section}.\arabic{subsection}.\arabic{subsubsection}}
\renewcommand{\theHfigure}{\Alph{figure}}
\renewcommand{\theHtable}{\Alph{table}}
\renewcommand{\theHtheorem}{\thesection.\arabic{theorem}}
\renewcommand{\theHequation}{\thesection.\arabic{equation}}

\definecolor{myorange}{RGB}{230,145,56}
\newcolumntype{Y}{>{\raggedleft\arraybackslash}X}
\newcommand{\ccell}[4]{#1/#2/#3/#4} 

\section*{Appendix}
\section{Algorithm}
Algorithm~\ref{alg:method} summarizes the complete MaPP training loop. At each step, prompts are first selected by sampling proportionally to the posterior-predictive score $V_G^{\mathrm{MaPP}}$ (\S\ref{sec:pp_selection}). After rollout, the posterior-predictive advantage $\hat{a}_j^{\mathrm{MaPP}}$ replaces the standard GRPO advantage for policy update (\S\ref{sec:recovery}). 
The shared Beta posterior is then updated from the new observations. The entire procedure adds only closed-form computations on top of standard GRPO, with no additional rollouts or model forward passes.

\begin{algorithm}[htbp]
\caption{\textsc{MaPP}: Marginalized Posterior-Predictive Framework}
\label{alg:method}
\begin{algorithmic}[1]
\Require Prompt pool $\mathcal{T}$; policy $\pi_\theta$; group size $G$; batch size $|\mathcal{B}|$; discount $\lambda \in (0,1)$; training steps $T$
\Ensure Updated policy $\pi_\theta$
\State Initialize $(\alpha_0^{\tau}, \beta_0^{\tau}) \leftarrow (1, 1)$ for all $\tau \in \mathcal{T}$ \Comment{uninformative Beta prior}
\For{$t = 0, \ldots, T-1$}
    \LeftComment{\textit{Posterior-predictive prompt selection (\S\ref{sec:pp_selection})}}
    \For{$\tau \in \mathcal{T}$}
        \State $V_G^{\mathrm{MaPP}}(\alpha_t^{\tau}, \beta_t^{\tau}) \gets 1 - \frac{B(\alpha_t^{\tau}+G,\, \beta_t^{\tau})}{B(\alpha_t^{\tau}, \beta_t^{\tau})} - \frac{B(\alpha_t^{\tau},\, \beta_t^{\tau}+G)}{B(\alpha_t^{\tau}, \beta_t^{\tau})}$ \Comment{Eq.~\eqref{eq:vg_pp}}
    \EndFor
    \State Sample batch $\mathcal{B}_t \sim q(\tau) \propto V_G^{\mathrm{MaPP}}(\alpha_t^{\tau}, \beta_t^{\tau})$
    \LeftComment{\textit{Rollout and reward collection}}
    \For{$\tau \in \mathcal{B}_t$}
        \State Draw $\{y_j^{\tau}\}_{j=1}^{G} \sim \pi_\theta(\cdot \mid \tau)$; observe rewards $\{r_{t;j}^{\tau}\}_{j=1}^G$
        \State $S_t^{\tau} \gets \textstyle\sum_{j=1}^{G} r_{t;j}^{\tau}$
    \EndFor
    \LeftComment{\textit{Posterior-predictive advantage (\S\ref{sec:recovery})}}
    \For{$\tau \in \mathcal{B}_t$, $j = 1, \ldots, G$}
        \State $\alpha_{t+1;j}'^{\tau} \gets \lambda \cdot \alpha_t^{\tau} + (1 - \lambda) \cdot \alpha_0^{\tau} + S_t^{\tau} - r_{t;j}^{\tau}$ \Comment{LOO posterior, Eq.~\eqref{eq:loo_posterior}}
        \State $\beta_{t+1;j}'^{\tau} \gets \lambda \cdot \beta_t^{\tau} + (1 - \lambda) \cdot \beta_0^{\tau} + G - S_t^{\tau} - 1 + r_{t;j}^{\tau}$
        \If{$r_{t;j}^{\tau} = 1$}
            \State $\hat{a}_{t;j}^{\mathrm{MaPP},\tau} \gets +\sum_{S=1}^{G-1} w_S(\alpha_{t+1;j}'^{\tau}, \beta_{t+1;j}'^{\tau}) \sqrt{(G-S)/S}$ \Comment{Eq.~\eqref{eq:pp_closed}}
        \Else
            \State $\hat{a}_{t;j}^{\mathrm{MaPP},\tau} \gets -\sum_{S=1}^{G-1} w_S(\beta_{t+1;j}'^{\tau}, \alpha_{t+1;j}'^{\tau}) \sqrt{(G-S)/S}$
        \EndIf
    \EndFor
    \LeftComment{\textit{Policy update (Eq.~\eqref{eq:final_grad})}}
    \State $\theta \gets \theta + \eta\, \nabla_\theta \mathcal{J}(\theta)$ \ \text{with} \ $\nabla_\theta \mathcal{J} = \frac{1}{|\mathcal{B}_t|}\!\sum_{\tau \in \mathcal{B}_t}\! \frac{1}{G}\!\sum_{j=1}^G \hat{a}_{t;j}^{\mathrm{MaPP},\tau} \nabla_\theta \log \pi_\theta(y_j^{\tau} \mid \tau)$
    \LeftComment{\textit{Posterior update (Eq.~\eqref{eq:beta_update})}}
    \For{$\tau \in \mathcal{B}_t$}
        \State $\alpha_{t+1}^{\tau} \gets \lambda \cdot \alpha_t^{\tau} + (1 - \lambda) \cdot \alpha_0^{\tau} + S_t^{\tau}$
        \State $\beta_{t+1}^{\tau} \gets \lambda \cdot \beta_t^{\tau} + (1 - \lambda) \cdot \beta_0^{\tau} + G - S_t^{\tau}$
    \EndFor
\EndFor
\end{algorithmic}
\end{algorithm}

\section{Implementation Details}
\label{app:implementation}

\subsection{Tasks and Datasets}

\textbf{Mathematics.}\quad We train on the MATH dataset~\citep{MATH}, which contains 7,500 competition-level problems spanning algebra, geometry, number theory, and combinatorics. Following prior work~\citep{dps,mopps,gps}, we evaluate on four benchmarks: AMC23, MATH500~\citep{MATH500}, Minerva Math~\citep{Minerva}, and OlympiadBench~\citep{Olym}, using the datasets hosted by MOPPS~\citep{mopps}. We adopt a binary reward function following the default configuration in verl~\citep{sheng2025hybridflow}: a reward of 1 for correct and 0 otherwise.

\textbf{Numerical planning.}\quad We adopt the Countdown Number Game~\citep{Countdown}, which requires combining given numbers using basic arithmetic operations to reach a target value. Training is conducted on a 2,000-problem subset of the Countdown-34 dataset. Evaluation uses two benchmarks: a 512-problem held-out split from Countdown-34 (CD-34), and a 512-problem subset from Countdown-4 (CD-4), a harder variant that provides four source numbers per problem and substantially enlarges the search space. 

\textbf{Visual geometry.}\quad We train on the 2,101-problem training split of the Geometry3k dataset~\citep{lu2021intergps,hiyouga2025geo3k}, which pairs geometric diagrams with multi-step reasoning questions. Evaluation is conducted on the official 601-problem test split. 



\subsection{Training Details}

We adopt GRPO~\citep{deepseekmath} as the default RLVR algorithm, implemented within the verl framework~\citep{sheng2025hybridflow}. 
At each training step, we sample $G{=}5$ responses per prompt for math tasks and $G{=}8$ responses per prompt for Countdown and Geometry, using temperature $1.0$ and top-$p{=}1.0$ for rollout generation.
We disable the KL penalty by setting $\beta{=}0$, consistent with~\citet{dapo}. Training batch sizes are set to 256 for MATH and Countdown, with mini-batch sizes of 128 and 64 respectively, and 512 for Geometry3k with a mini-batch size of 256. The maximum response length is 1024 tokens for all tasks. Optimization is performed using AdamW~\citep{loshchilov2017decoupled} with learning rate $1 \times 10^{-6}$, $(\beta_1, \beta_2) = (0.9, 0.999)$, and weight decay 0.01. We apply the Clip-Higher strategy from DAPO~\citep{dapo}, which decouples clipping ranges with $\epsilon_{\mathrm{low}}{=}0.2$ and $\epsilon_{\mathrm{high}}{=}0.28$. All experiments are conducted on 8 NVIDIA H100 GPUs.

For all prediction-based prompt selection methods, the candidate pool size is set to $\hat{M}{=}8\times$ the training batch size $B$. 
For MoPPS~\citep{mopps} and DPS~\citep{dps}, we follow their original top-$B$ selection protocol, where prompts are ranked by the corresponding scoring criterion and the top-$B$ prompts are selected. 
In contrast, MaPP does not perform deterministic top-$B$ selection. Instead, each candidate prompt is assigned a retention probability $q(\tau) \propto V_G^{\mathrm{MaPP}}(\alpha^\tau,\beta^\tau)$, and the training batch is sampled according to these probabilities. 
For MaPP, the Beta prior is initialized as $(\alpha_0, \beta_0) = (1, 1)$ and the temporal discount is set to $\lambda{=}0.5$ across all tasks. 
For MoPPS~\citep{mopps}, we follow the original configuration with Beta prior $(\alpha_0, \beta_0) = (1, 1)$, target success probability $\gamma^{*}{=}0.5$, and decay factor $\lambda{=}0.5$. 
For DPS~\citep{dps}, we adopt the default three-state HMM with Dirichlet prior $\alpha_0 = (1, 1, 1)$ and decay ratio $\lambda{=}0.5$. 
For Dynamic Sampling~\citep{dapo}, we use the verl implementation, which over-samples a candidate batch $4\times$ the training batch size and post-hoc filters out prompts with zero reward standard deviation.

\section{Additional Experiments}
\label{app:additional_exp}


\subsection{Empirical Calibration of MaPP-AD and MaPP-PS}
\label{app:calibration}
\begin{figure*}[t]
    \centering
    \includegraphics[width=\textwidth]{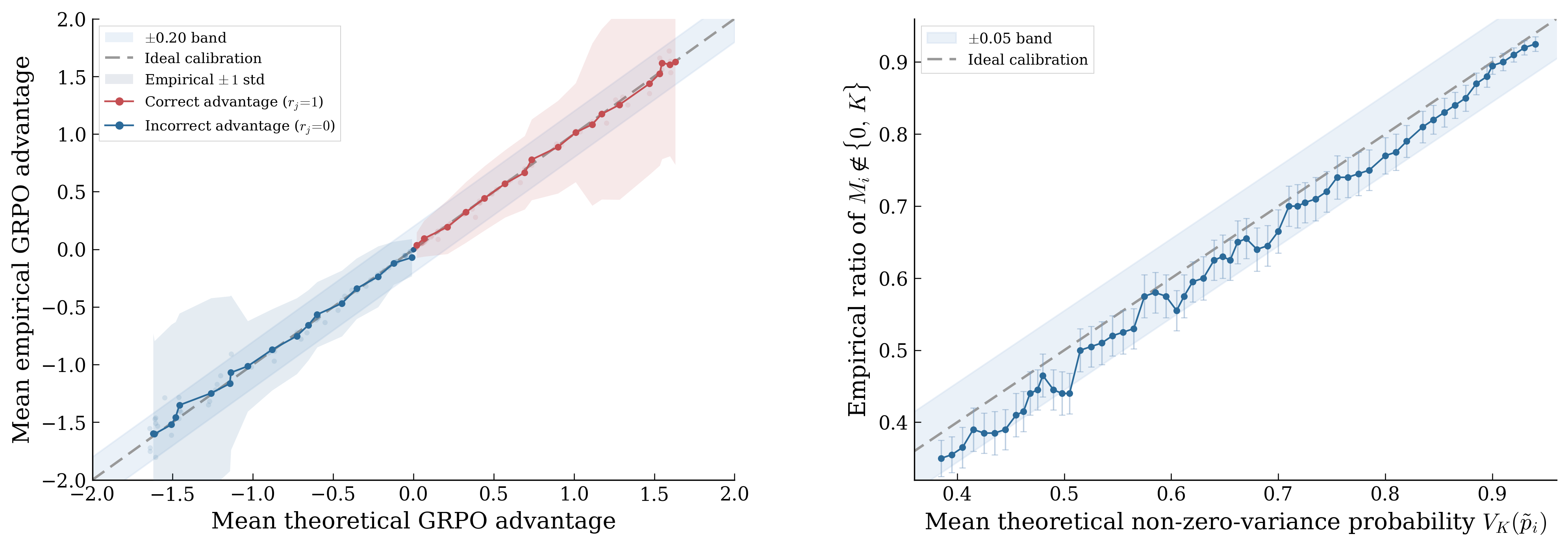}
    \caption{\textbf{(a)} Calibration of the intrinsic advantage: the theoretical prediction $a_j^{\star}(\gamma)$ (Proposition~\ref{prop:intrinsic}) closely matches the empirical mean advantage across 100 prompts $\times$ 50 groups ($G{=}8$). The $\pm 1$ std band reflects the composition noise that MaPP eliminates via marginalization.
\textbf{(b)} Calibration of the selection score: the posterior-predictive non-zero-variance probability $V_G^{\mathrm{MaPP}}$ (Eq.~\eqref{eq:vg_pp}) tracks the empirical effective ratio within a $\pm 0.05$ band.}
\label{fig:advantage_noise}
\end{figure*}

We verify that $a_j^{\star}(\gamma)$ and $V_G^{MaPP}(\gamma)$ faithfully reflect empirical training dynamics. 
For MaPP-AD, we take a mid-training checkpoint on Countdown (Qwen3-4B, $G{=}8$), sample 100 prompts and draw 50 independent groups per prompt; Figure~\ref{fig:advantage_noise}(a) shows that the theoretical intrinsic advantage closely tracks the empirical mean advantage, with nearly all points falling along the ideal calibration line, and the $\pm 1$ std band visualizes the composition noise that MaPP-AD eliminates. 
For MaPP-PS, we collect the predicted $V_G^{MaPP}(\hat{\gamma})$ and the empirical non-degenerate ratio at each training step throughout training, bin them by the theoretical score, and compare; Figure~\ref{fig:advantage_noise}(b) shows that $V_G(\hat{\gamma})$ tracks the empirical ratio within a $\pm 0.05$ band across the full range, confirming reliable difficulty estimation. Together, both plots validate the Bernoulli--Beta modeling assumption that MaPP builds upon.

\subsection{Training Dynamics}
\label{app:training_dynamics}

\begin{figure*}[t]
    \centering
    \includegraphics[width=\textwidth]{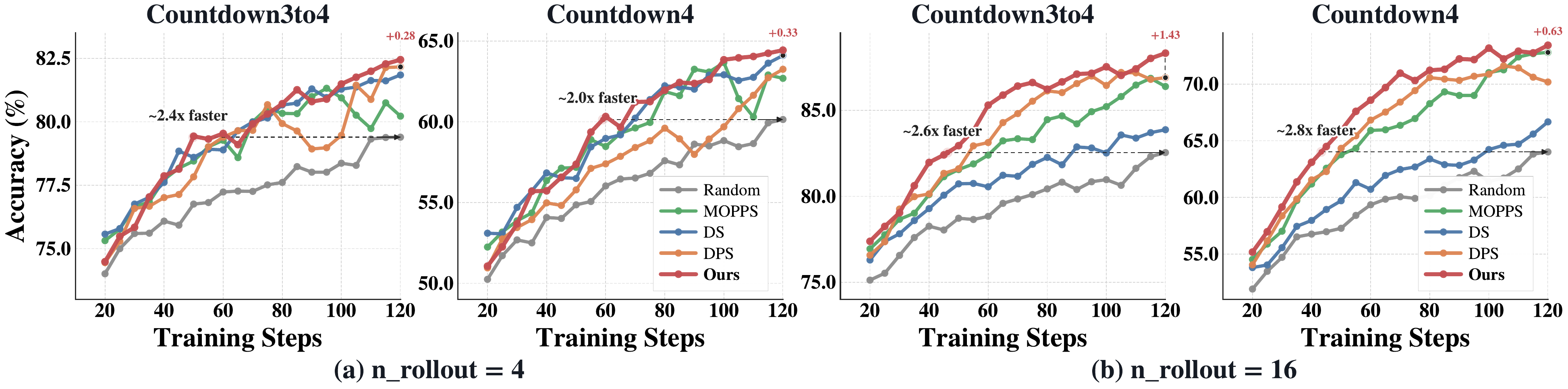}
    \caption{Ablation experiments of our proposed MaPP method under different numbers of rollouts ($n=4$ and $n=16$, with $n=8$ evaluated in the main experiments). The results show that {MaPP} performs consistently well across different rollout group sizes.}
    \label{fig:abtion_group}
    
\end{figure*}

\paragraph{Robustness to group size.} Figure~\ref{fig:abtion_group} shows accuracy versus training steps under group sizes $G{=}4$ and $G{=}16$ on Countdown. MaPP reaches the same accuracy as the strongest baseline by approximately $2.0$--$2.4\times$ faster convergence speed at $G{=}4$, while the acceleration increases to $2.6$--$2.8\times$ at $G{=}16$. The more pronounced speedup at larger $G$ is consistent with the observation in \S\ref{sec:ablation} that richer per-step evidence enables sharper advantage corrections.

\begin{figure*}[t]
    \centering
    \includegraphics[width=\textwidth]{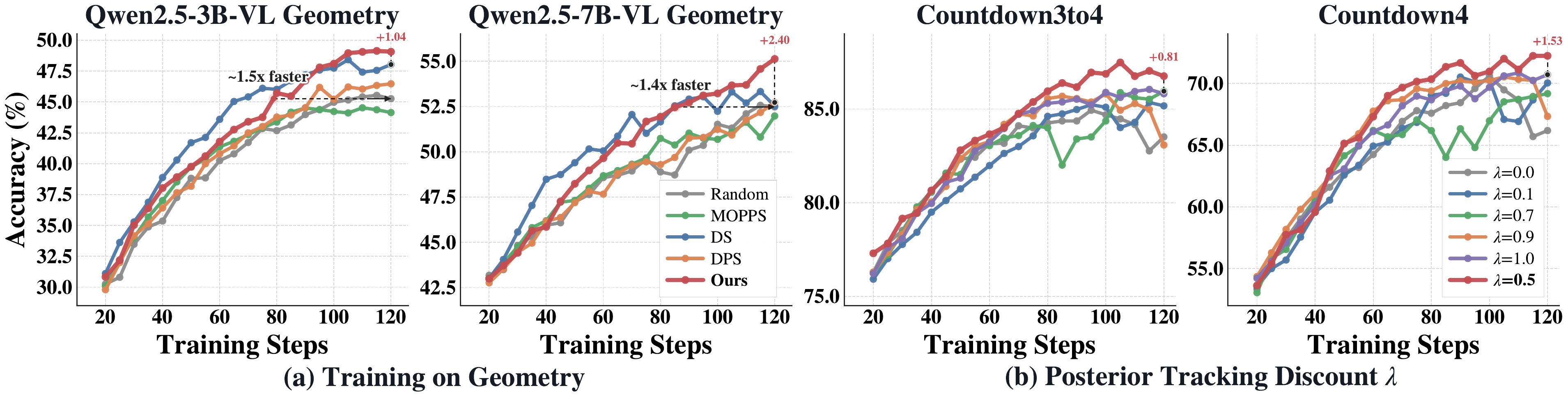}
    \caption{(a): Performance of our method trained on the Qwen2.5-VL-Instruct series models and evaluated on the Geometry test set. (b): Self-ablation of MaPP under different posterior tracking discounts. The best performance is achieved at $\lambda=0.5$ (Ours).}
    \label{fig:abtion_lambda}

\end{figure*}

\paragraph{Visual geometry and sensitivity to $\lambda$.} Figure~\ref{fig:abtion_lambda}(a) shows training curves for the visual geometry task on Qwen2.5-VL-3B and Qwen2.5-VL-7B. MaPP outperforms the strongest baseline by +1.04 and +2.40 respectively at convergence, reaching comparable accuracy by approximately $1.4$--$1.5\times$ faster convergence speed, confirming that MaPP generalizes to multi-modal settings. Figure~\ref{fig:abtion_lambda}(b) shows training curves under varying $\lambda$ on Countdown. $\lambda{=}0.5$ achieves the best final accuracy and convergence speed, while extreme values ($\lambda{=}0.0$ and $\lambda{=}1.0$) converge noticeably slower.

\section{Theoretical Analysis}
\label{app:proofs}

All proofs in this section are for a single prompt $\tau$ at a fixed training step $t$; we drop the superscript $\tau$ for brevity and write $\alpha_j' := \alpha_{t+1;j}'$, $\beta_j' := \beta_{t+1;j}'$ following Proposition~\ref{prop:pp_closed}. The latent pass rate $\gamma\in(0,1)$ is treated as a fixed (but unknown) parameter; expectations $\mathbb{E}[\cdot\mid\gamma]$ are taken over the rollout randomness only, while Bayesian quantities are computed under the leave-one-out posterior $\gamma\mid\mathcal{H}_t,r_{t;-j}\sim\mathrm{Beta}(\alpha_j',\beta_j')$. Under binary rewards, the GRPO advantage reduces to
\begin{equation}
\hat{A}_j = h(r_j, S) = \begin{cases} +\!\sqrt{(G-S)/S}, & r_j=1,\\ -\!\sqrt{S/(G-S)}, & r_j=0, \end{cases}
\label{eq:app:h}
\end{equation}
with $\hat{A}_j:=0$ when $S \in \{0,G\}$; correspondingly, both $a_j^{\star}(\gamma)$ and $\hat{A}_j$ drop the degenerate terms, so $\mathbb{E}[\hat{A}_j\mid r_j,\gamma] = a_j^{\star}(\gamma)$ for $\gamma\in(0,1)$.

\begin{theorem}[Composition-noise lower bound]
\label{thm:floor}
For any $G\ge 2$ and $\gamma\in(0,1)$,
\begin{equation}
\mathbb{E}\!\big[(\hat{A}_j-a_j^{\star}(\gamma))^2\,\big|\,\gamma\big] \;=\; \xi(\gamma) \;>\; 0,
\label{eq:app:floor}
\end{equation}
where $\xi(\gamma)$ depends only on $(\gamma,G)$ and is independent of the posterior or training history.
\end{theorem}

\paragraph{Proof.} By Definition~\ref{def:intrinsic}, $a_j^{\star}(\gamma) = \mathbb{E}[\hat{A}_j\mid r_j,\gamma]$. Conditioning on $r_j$ and $\gamma$, the only remaining randomness comes from the other $G-1$ responses, so
\begin{equation}
\mathbb{E}\!\big[(\hat{A}_j - a_j^{\star}(\gamma))^2 \mid \gamma\big] = \mathbb{E}_{r_j}\!\big[\operatorname{Var}(\hat{A}_j \mid r_j, \gamma)\,\big|\,\gamma\big] =: \xi(\gamma).
\end{equation}
Letting $k\sim\mathrm{Bin}(G{-}1,\gamma)$ denote the number of correct responses among the other $G-1$,
\begin{equation}
\xi(\gamma) = \gamma\,\operatorname{Var}_{k}\!\!\left[\sqrt{\tfrac{G{-}1{-}k}{k{+}1}}\right] + (1{-}\gamma)\,\operatorname{Var}_{k}\!\!\left[\sqrt{\tfrac{k}{G{-}k}}\right].
\label{eq:app:RG_explicit}
\end{equation}
The map $S\mapsto\sqrt{(G-S)/S}$ is strictly monotone on $\{1,\dots,G-1\}$ and the binomial is non-degenerate for $\gamma\in(0,1)$, hence $\xi(\gamma)>0$. Since $\xi$ depends only on $(\gamma,G)$, this lower bound is structural and cannot be reduced by any prompt-level posterior or selection scheme. \hfill$\square$

\begin{theorem}[Frequentist MSE bound for MaPP]
\label{thm:freq}
Fix $\gamma\in(0,1)$ and $G\ge 2$. Let $\tilde\gamma:=\alpha_j'/(\alpha_j'+\beta_j')$ and $n:=\alpha_j'+\beta_j'$. There exist constants $C_1(\gamma,G), C_2(\gamma,G)$ depending only on $(\gamma,G)$ such that
\begin{equation}
\mathbb{E}\!\big[(\hat{a}_j^{\mathrm{MaPP}} - a_j^{\star}(\gamma))^2\,\big|\,\gamma\big] \;\le\; \underbrace{C_1\,(\tilde\gamma-\gamma)^2}_{\text{calibration bias}} \;+\; \underbrace{\frac{C_2}{n+1}}_{=\,O(G/n)}.
\label{eq:app:freq_bound}
\end{equation}
\end{theorem}
\paragraph{Proof.} Decompose the frequentist MSE under the rollout history $\mathcal{H}_t$ into squared bias and variance:
\begin{equation}
\mathbb{E}\!\big[(\hat{a}_j^{\mathrm{MaPP}}-a_j^{\star}(\gamma))^2\mid\gamma\big] \;=\; \underbrace{\big(\mathbb{E}_{\mathcal{H}_t}[\hat{a}_j^{\mathrm{MaPP}}\mid\gamma]-a_j^{\star}(\gamma)\big)^2}_{\text{bias}^2} \;+\; \underbrace{\operatorname{Var}_{\mathcal{H}_t}(\hat{a}_j^{\mathrm{MaPP}}\mid\gamma)}_{\text{variance}}.
\label{eq:app:bias_var}
\end{equation}
Since $a_j^{\star}=\pm\mu_G$ is a degree-$(G{-}1)$ polynomial with binomial coefficients, its first two derivatives are bounded on compact subsets of $(0,1)$:
\begin{equation}
L_G \;:=\; \sup_{\gamma\in K}\,\big|(a_j^{\star})'(\gamma)\big| \;=\; O(G),\qquad M_G \;:=\; \sup_{\gamma\in K}\,\big|(a_j^{\star})''(\gamma)\big| \;=\; O(G^2),\qquad K\Subset(0,1),
\label{eq:app:lipschitz}
\end{equation}
where the constants depend on the distance of $K$ to $\{0,1\}$ and are bounded over the non-degenerate regime enforced by MaPP-PS via $V_G^{\mathrm{MaPP}}$.
A second-order Taylor expansion of $a_j^{\star}$ around $\gamma$ gives, for $\gamma'$ in a neighborhood of $\gamma$,
\begin{equation}
a_j^{\star}(\gamma') \;=\; a_j^{\star}(\gamma) + (a_j^{\star})'(\gamma)\,(\gamma'-\gamma) + R(\gamma',\gamma),\qquad |R(\gamma',\gamma)|\le \tfrac{1}{2}M_G\,(\gamma'-\gamma)^2.
\label{eq:app:taylor}
\end{equation}
Taking expectation under the leave-one-out posterior $\gamma'\sim\mathrm{Beta}(\alpha_j',\beta_j')$ at fixed $\mathcal{H}_t$, and using $\mathbb{E}_{\gamma'}[\gamma']=\tilde\gamma$,
\begin{equation}
\hat{a}_j^{\mathrm{MaPP}} \;=\; a_j^{\star}(\gamma) + (a_j^{\star})'(\gamma)\,(\tilde\gamma-\gamma) + \mathbb{E}_{\gamma'}[R(\gamma',\gamma)].
\label{eq:app:mapp_expansion}
\end{equation}

Bias. 
Taking expectation of Eq.~\eqref{eq:app:mapp_expansion} over $\mathcal{H}_t$ at fixed $\gamma$ and using $|\mathbb{E}_{\gamma'}[R]|\le \tfrac{1}{2}M_G\mathbb{E}_{\gamma'}[(\gamma'-\gamma)^2]$,
\begin{equation}
\big|\mathbb{E}_{\mathcal{H}_t}[\hat{a}_j^{\mathrm{MaPP}}\mid\gamma] - a_j^{\star}(\gamma)\big| \;\le\; L_G\,\big|\mathbb{E}_{\mathcal{H}_t}[\tilde\gamma\mid\gamma]-\gamma\big| + O\!\big(M_G/n\big),
\end{equation}
where the last term absorbs both $\mathbb{E}_{\gamma'}[(\gamma'-\tilde\gamma)^2]=\operatorname{Var}_{\gamma'}(\gamma')=O(1/n)$ and $(\tilde\gamma-\gamma)^2$. Squaring, the bias contributes $C_1\,(\tilde\gamma-\gamma)^2 + O(G^2/n^2)$ with $C_1 = L_G^2 = O(G^2)$; the higher-order term is dominated by the variance term below for $n\ge 1$.

Variance. 
The variance term equals the posterior variance of $a_j^{\star}(\gamma')$ under $\gamma'\sim\mathrm{Beta}(\alpha_j',\beta_j')$. Substituting Eq.~\eqref{eq:app:taylor} and using $\operatorname{Var}_{\gamma'}(\text{constants})=0$,
\begin{equation}
\operatorname{Var}_{\gamma'}\!\big(a_j^{\star}(\gamma')\big) \;\le\; \big[(a_j^{\star})'(\gamma)\big]^2\operatorname{Var}_{\gamma'}(\gamma') + O\!\big(M_G^2/n^2\big) \;\le\; \frac{L_G^2}{4(n+1)} + O(1/n^2),
\end{equation}
using $\operatorname{Var}_{\gamma'}(\gamma')=\alpha_j'\beta_j'/[n^2(n+1)]\le 1/[4(n+1)]$. This gives $C_2 = L_G^2/4 = O(G)$.

Combining bias and variance yields Eq.~\eqref{eq:app:freq_bound}.


\begin{corollary}[Crossover]
\label{cor:safety}
Combining Theorems~\ref{thm:floor} and~\ref{thm:freq}, there exists a threshold
\begin{equation}
n^{\star}(\gamma,G) \;\le\; \frac{C_2}{\xi(\gamma)-C_1(\tilde\gamma-\gamma)^2}-1,
\end{equation}
such that for all $n\ge n^{\star}$,
\begin{equation}
\mathbb{E}\!\big[(\hat{a}_j^{\mathrm{MaPP}} - a_j^{\star}(\gamma))^2\,\big|\,\gamma\big]\;\le\;\xi(\gamma) \;=\; \mathbb{E}\!\big[(\hat{A}_j-a_j^{\star}(\gamma))^2\,\big|\,\gamma\big],
\end{equation}
i.e., MaPP strictly improves over GRPO at the per-response level. In practice, this threshold is small under standard initialization $(\alpha_0,\beta_0)=(1,1)$ and is reached after a few posterior updates per prompt, ensuring that MaPP outperforms GRPO throughout the main training phase.
\end{corollary}


\paragraph{Connection to gradient estimation.} The per-response MSE bounds in Theorems~\ref{thm:floor} and~\ref{thm:freq} transfer to the policy gradient under a standard score-function moment condition. Let $s_j := \nabla_\theta \log \pi_\theta(y_j \mid \tau)$ and let $g := \frac{1}{G}\sum_{j=1}^{G} \hat{a}_j s_j$, $g^{\star} := \frac{1}{G}\sum_{j=1}^{G} a_j^{\star}(\gamma) s_j$ denote the policy gradient and its oracle counterpart, where $\hat{a}_j$ denotes either $\hat{A}_j$ (GRPO) or $\hat{a}_j^{\mathrm{MaPP}}$ (MaPP), with per-response error $\epsilon_j := \hat{a}_j - a_j^{\star}(\gamma)$. Suppose $\mathbb{E}[\|s_j\|^2 \mid \tau] \le B^2$, as is routinely assumed in policy-gradient analysis. A direct application of Cauchy--Schwarz to $g - g^{\star} = \frac{1}{G}\sum_{j=1}^{G} \epsilon_j s_j$, together with symmetry across responses, yields
\begin{equation}
\mathbb{E}\!\big[\|g - g^{\star}\|^2 \,\big|\, \gamma\big] \;\le\; B^2 \cdot \mathbb{E}\!\big[\epsilon_1^2 \,\big|\, \gamma\big].
\end{equation}
Since this bound is monotone in the per-response MSE, the bounds of Theorems~\ref{thm:floor} and~\ref{thm:freq} immediately yield corresponding gradient-level bounds, and Corollary~\ref{cor:safety} likewise transfers: MaPP achieves a smaller gradient MSE upper bound than GRPO whenever $n \ge n^{\star}$. Empirically, Fig.~\ref{fig:teaser}(b) and Fig.~\ref{fig:advantage_noise} show that composition noise produces visible variability in realized advantages across groups, which MaPP suppresses.

\section{Derivation of Closed Forms}
\label{app:closed_forms}

All derivations in this section are for a single prompt $\tau$; we drop the superscript $\tau$ for brevity and write $\alpha_j' := \alpha_{t+1;j}'$, $\beta_j' := \beta_{t+1;j}'$ following Proposition~\ref{prop:pp_closed}.

\paragraph{Intrinsic advantage (Proposition~\ref{prop:intrinsic}).} Consider $r_j=1$; the case $r_j=0$ follows by symmetry. Conditional on $r_j=1$ and $\gamma$, the remaining $G-1$ responses each follow independent $\mathrm{Bernoulli}(\gamma)$ trials, so the total success count satisfies $S \in \{1, \ldots, G\}$ with $\mathbb{P}(S=s \mid r_j=1, \gamma) = \binom{G-1}{s-1}\gamma^{s-1}(1-\gamma)^{G-s}$. Since $\hat{A}_j=0$ when $S=G$, taking expectation gives
\begin{align}
a_j^{\star}(\gamma) &= \mathbb{E}\!\big[\hat{A}_j \mid r_j{=}1,\,\gamma\big] = \sum_{S=1}^{G-1} \binom{G-1}{S-1}\gamma^{S-1}(1{-}\gamma)^{G-S} \sqrt{\frac{G{-}S}{S}} = \mu_G(\gamma).
\label{eq:app:intrinsic}
\end{align}
For $r_j=0$, by symmetry $a_j^{\star}(\gamma)=-\mu_G(1{-}\gamma)$.

\paragraph{Posterior-predictive advantage (Proposition~\ref{prop:pp_closed}).} Marginalizing $a_j^{\star}(\gamma)=\mu_G(\gamma)$ against the leave-one-out posterior $\gamma\sim\mathrm{Beta}(\alpha_j',\beta_j')$ and exchanging summation with integration:
\begin{align}
\hat{a}_j^{\mathrm{MaPP}} &= \int_0^1 \mu_G(\gamma)\; \frac{\gamma^{\alpha_j'-1}(1{-}\gamma)^{\beta_j'-1}} {B(\alpha_j',\beta_j')}\,\mathrm{d}\gamma \notag\\ &= \sum_{S=1}^{G-1} \binom{G{-}1}{S{-}1} \sqrt{\frac{G{-}S}{S}}\; \frac{1}{B(\alpha_j',\beta_j')} \underbrace{ \int_0^1 \gamma^{\alpha_j'+S-2} (1{-}\gamma)^{\beta_j'+G-S-1}\,\mathrm{d}\gamma }_{=\,B(\alpha_j'+S-1,\;\beta_j'+G-S)} \notag\\ &= \sum_{S=1}^{G-1} \underbrace{\binom{G{-}1}{S{-}1} \frac{B(\alpha_j'{+}S{-}1,\;\beta_j'{+}G{-}S)} {B(\alpha_j',\beta_j')}}_{=\;w_S(\alpha_j',\,\beta_j')\; =\;P_{\mathrm{BB}}(S{-}1\mid G{-}1,\,\alpha_j',\beta_j')} \;\sqrt{\frac{G{-}S}{S}}.
\label{eq:app:pp}
\end{align}
The case $r_j=0$ follows by the change of variable $\gamma'=1{-}\gamma$, which swaps $\alpha_j'$ and $\beta_j'$.


\paragraph{Selection score (Eq.~\eqref{eq:vg_pp}).} 
The selection score operates across prompts, so we restore the superscript $\tau$. By linearity, 
$V_G^{\mathrm{MaPP}} = 1-\mathbb{E}_{\gamma\sim\mathrm{Beta}(\alpha^{\tau}, \beta^{\tau})}[\gamma^G] - \mathbb{E}[(1{-}\gamma)^G]$. 
The first moment evaluates as
\begin{equation}
\mathbb{E}[\gamma^G] = \frac{1}{B(\alpha^{\tau}, \beta^{\tau})} \int_0^1 \gamma^{\alpha^{\tau}+G-1} (1{-}\gamma)^{\beta^{\tau}-1}\,\mathrm{d}\gamma = \frac{B(\alpha^{\tau}{+}G,\;\beta^{\tau})} {B(\alpha^{\tau},\beta^{\tau})},
\end{equation}
and $\mathbb{E}[(1{-}\gamma)^G]$ follows by symmetry, giving Eq.~\eqref{eq:vg_pp}.

\section{Extension to Multi-Level Rewards}
\label{app:multilevel}
The main text presents MaPP under binary rewards for notational clarity, but the posterior-predictive marginalization principle extends naturally to discrete multi-level rewards $r_j^{\tau} \in \{v_1, \ldots, v_L\}$ by replacing the Beta--Binomial conjugate pair with the Dirichlet--Multinomial pair: the latent difficulty becomes a probability vector $\boldsymbol{\gamma}^{\tau} \in \Delta^{L-1}$ with a Dirichlet
posterior, and the intrinsic advantage is obtained by enumerating over leave-one-out count vectors weighted by the
Dirichlet--Multinomial predictive probability. For moderate $G$
and $L$, this involves $\binom{G+L-2}{L-1}$ terms (e.g., 36 for
$G{=}8$, $L{=}3$), which is computationally negligible.
In this work, all experiments are conducted under binary rewards:
the standard $\{0,1\}$ setting is used directly, and the
$\{0,\,0.1,\,1\}$ reward scheme is also treated as binary by
merging $0$ and $0.1$ into the negative class, following
MOPPS~\citep{mopps} and related methods.
A systematic evaluation of the multi-level extension is left for
future work.

\section{Data Examples}

We provide illustrative data examples for each task below. Prompt templates for MATH and Geometry3k are adopted from the verl framework \citep{sheng2025hybridflow}, and the Countdown template follows \citet{Countdown}.

\newpage
\newtcolorbox{examplebox}[1]{
    enhanced,
    breakable,
    colback=white,
    colframe=black,
    colbacktitle=white,
    coltitle=black,
    fonttitle=\bfseries\normalsize,
    title={#1},
    boxrule=0.5pt,
    sharp corners,
    left=12pt,
    right=12pt,
    top=10pt,
    bottom=10pt,
    toptitle=2pt,
    bottomtitle=2pt,
    titlerule=0.5pt
}

\begin{examplebox}{MATH Data Example}
\textbf{Prompt:}

\vspace{4pt}

You have seven bags of gold coins. Each bag has the same number of gold coins. One day, you find a bag of $53$ coins. You decide to redistribute the number of coins you have so that all eight bags you hold have the same number of coins. You successfully manage to redistribute all the coins, and you also note that you have more than $200$ coins. What is the smallest number of coins you could have had before finding the bag of $53$ coins?

\vspace{4pt}

Let's think step by step and output the final answer within \verb|\boxed{}|.

\vspace{6pt}

\textbf{Ground-Truth Answer:}

\vspace{2pt}

203
\end{examplebox}

\vspace{8pt}

\begin{examplebox}{Countdown Data Example}
\textbf{Prompt:}

\vspace{4pt}

A conversation between User and Assistant. The user asks a question, and the Assistant solves it.  
The assistant first thinks about the reasoning process in the mind and then provides the user with the answer.

\vspace{3pt}

User: Using the numbers $[79, 8, 27, 47]$, create an equation that equals $91$.  
You can use basic arithmetic operations $(+, -, \times, \div)$, and each number can only be used once.  
Show your work in \texttt{<think>} \texttt{</think>} tags, and return the final answer in \texttt{<answer>} \texttt{</answer>} tags, for example \texttt{<answer>}$(1+2)/3$\texttt{</answer>}.

\vspace{3pt}

Assistant: Let me solve this step by step.

\texttt{<think>}

\vspace{6pt}

\textbf{Ground-Truth Answer:}

\vspace{2pt}

$79 + 47 - 27 - 8 = 91$
\end{examplebox}

\vspace{8pt}

\begin{examplebox}{Geometry3K Data Example}
\textbf{Prompt:}

\vspace{4pt}

\begin{minipage}{0.34\linewidth}
    \centering
    \includegraphics[width=0.95\linewidth]{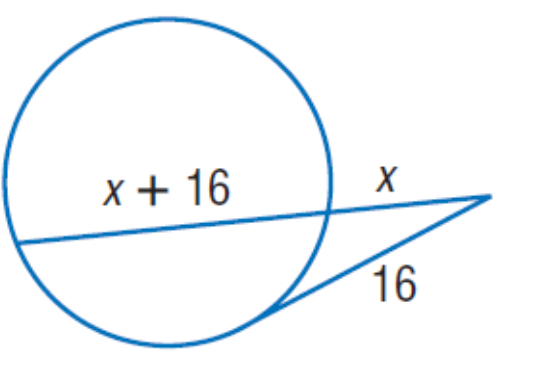}
\end{minipage}
\hfill
\begin{minipage}{0.62\linewidth}
Find $x$. Round to the nearest tenth if necessary. Assume that segments that appear to be tangent are tangent.

\vspace{4pt}

You FIRST think about the reasoning process as an internal monologue and then provide the final answer.  
The reasoning process MUST BE enclosed within \texttt{<think>} \texttt{</think>} tags.  
The final answer MUST BE put in \verb|\boxed{}|.
\end{minipage}

\vspace{6pt}

\textbf{Ground-Truth Answer:}

\vspace{2pt}

8
\end{examplebox}

\end{document}